\documentclass[10pt, a4paper]{article}
\usepackage{lrec2022} 
\usepackage{multibib}
\newcites{languageresource}{Language Resources}
\usepackage{graphicx}
\usepackage{tabularx}
\usepackage{soul}
\usepackage{titlesec}
\titleformat{\section}{\normalfont\large\bf\center}{\thesection.}{1em}{}
\titleformat{\subsection}{\normalfont\SmallTitleFont\bf\raggedright}{\thesubsection.}{1em}{}
\titleformat{\subsubsection}{\normalfont\normalsize\bf\raggedright}{\thesubsubsection.}{1em}{}
\renewcommand\thesection{\arabic{section}}
\renewcommand\thesubsection{\thesection.\arabic{subsection}}
\renewcommand\thesubsubsection{\thesubsection.\arabic{subsubsection}}

\usepackage{epstopdf}
\usepackage[utf8]{inputenc}

\usepackage{hyperref}
\usepackage{xstring}

\usepackage{color}

\title{Distilling the Knowledge of Romanian BERTs Using Multiple Teachers}

\name{Andrei-Marius Avram\textsuperscript{1}, Darius Catrina\textsuperscript{*}\thanks{\textsuperscript{*}Work done during an interhsip at the Research Institute for Artificial Intelligence, Romanian Academy.}\textsuperscript{3}, Dumitru-Clementin Cercel\textsuperscript{2}, Mihai Dascalu\textsuperscript{2},\\
\large{\textbf{Traian Rebedea\textsuperscript{2}, Vasile Păiș\textsuperscript{1}, Dan Tufiș\textsuperscript{1}}}}

\address{Research Institute for Artificial Intelligence, Romanian Academy\textsuperscript{1} \\
University Politehnica of Bucharest, Faculty of Automatic Control and Computers\textsuperscript{2} \\
Duke University\textsuperscript{3}\\
         \{andrei.avram, vasile, tufis\}@racai.ro, \{dumitru.cercel, mihai.dascalu, traian.rebedea\}@upb.ro\\}

\abstract{
Running large-scale pre-trained language models in computationally constrained environments remains a challenging problem yet to be addressed, while transfer learning from these models has become prevalent in Natural Language Processing tasks. Several solutions, including knowledge distillation, network quantization, or network pruning have been previously proposed; however, these approaches focus mostly on the English language, thus widening the gap when considering low-resource languages. In this work, we introduce three light and fast versions of distilled BERT models for the Romanian language: Distil-BERT-base-ro, Distil-RoBERT-base, and DistilMulti-BERT-base-ro. The first two models resulted from the individual distillation of knowledge from two base versions of Romanian BERTs available in literature, while the last one was obtained by distilling their ensemble. To our knowledge, this is the first attempt to create publicly available Romanian distilled BERT models, which were thoroughly evaluated on five tasks: part-of-speech tagging, named entity recognition, sentiment analysis, semantic textual similarity, and dialect identification. Our experimental results argue that the three distilled models offer performance comparable to their teachers, while being twice as fast on a GPU and $\sim$35\% smaller. In addition, we further test the similarity between the predictions of our students versus their teachers by measuring their label and probability loyalty, together with regression loyalty - a new metric introduced in this work.
 \\ \newline \Keywords{Knowledge Distillation, BERT, Romanian language, Multiple Teachers, Loyalty} }

\begin{document}

\maketitleabstract

\section{Introduction}

Knowledge transfer from Transformer-based language models \cite{vaswani2017attention} trained on large amounts of data achieves state-of-the-art results on most Natural Language Processing (NLP) tasks \cite{devlin2019bert,liu2019roberta,he2020deberta}. However, the best performing models usually have billions \cite{brown2020language} or even trillions \cite{fedus2021switch} of parameters, making them impractical in certain real-world situations. Moreover, both training and using these language models usually comes at a high environmental cost \cite{strubell2019energy}.

Several attempts were made to reduce the size of models by distilling their knowledge \cite{44873} accumulated during the pre-training phase \cite{sanh2019distilbert}, after fine-tuning the model on a specific task \cite{turc2019well}, or both pre-training and fine-tuning \cite{jiao2020tinybert}. Other methods consider shrinking the size of the models by either quantizing their weights to integer values \cite{shen2020q}, or pruning parts of the neural network \cite{brix2020successfully}. In addition, efficient attention mechanisms were developed to overcome the quadratic bottleneck in the sequence length of multi-head attention \cite{zaheer2020big,choromanski2020rethinking}.

However, the vast majority of these efforts focused on developing English models, and little attention was paid on increasing the efficiency of pre-trained models on other languages, with few singular cases of such compressed models like BERTino \cite{muffobertino} for Italian, MBERTA for Arabic \cite{alyafeai2021arabic}, or GermDistilBERT\footnote{\url{https://huggingface.co/distilbert-base-german-cased}} for German. As a response to this issue, we focus on Romanian, a language on which BERT has recently attracted a surge of attention from the local community and has shown promising results in various areas like dialect identification \cite{zaharia2021dialect,popa2020applying,zaharia2020exploring}, document classification \cite{avram2021pyeurovoc}  or satire detection in news \cite{rogoz2021saroco}. Thus, our work introduces three compressed BERT versions for the Romanian language that were obtained through a distillation process:
\begin{itemize}
    \item \textbf{Distil-BERT-base-ro}\footnote{\url{https://huggingface.co/racai/distilbert-base-romanian-cased}} was obtained by distilling the knowledge of BERT-base-ro \cite{dumitrescu2020birth} using its original training corpus and tokenizer;
    \item \textbf{Distil-RoBERT-base}\footnote{\url{https://huggingface.co/racai/distilbert-base-romanian-uncased}} was created from RoBERT-base \cite{masala2020robert} in similar conditions (i.e., using both original training corpus and tokenizer);
    \item \textbf{DistilMulti-BERT-base-ro}\footnote{\url{https://huggingface.co/racai/distilbert-multi-base-romanian-cased}} considered the distillation of the knowledge from an ensemble consisting of BERT-base-ro and RoBERT-base, while relying on the combined corpus and coupled with the tokenizer of the former model.
\end{itemize}

Our three compressed models were further evaluated on five Romanian datasets and the results showed that they maintained most of the performance of the original models, while being approximately twice as fast when run on a GPU. In addition, we also measure the label, probability and regression loyalties between each of the three distilled models and their teachers, as quantifying the performance on specific tasks does not show how similar the predictions between a teacher and a student really are. The models, together with the distillation and evaluation scripts were open-sourced to improve the reproducibility of this work\footnote{\url{https://github.com/racai-ai/Romanian-DistilBERT}}.

The rest of the paper is structured as follows. The next section presents a series of solutions related to the knowledge distillation of pre-trained language models. The third section outlines our approach of distilling the knowledge of Romanian BERTs, whereas the fourth section presents the evaluation setup and their performance on various Romanian tasks. The fifth section evaluates the prediction loyalty between each distilled version and its teacher, while the sixth section evaluates their inference speed. The final section concludes our work and outlines potential future work.

\section{Related Work}

Knowledge Distillation \cite{44873} is a compression method in which a smaller framework, the student model, is trained to reduce the loss $\mathcal{L}_{KD}$ over the soft probabilities predicted by a larger model (i.e., the teacher):
\begin{equation}
    \mathcal{L}_{KD} = \sum_i t_i \cdot log (s_i)
    \label{eq:distil}
\end{equation}
where $t_i$ and $s_i$ are the probabilities predicted by the teacher and the student, respectively.

The technique usually uses a temperature parameter $T$ in the softmax function that controls the smoothness of the distribution given by probabilities $p_i$, known as softmax-temperature:
\begin{equation}
    p_i = \frac{exp(z_i / T)}{\sum_j exp(z_j / T)}
    \label{eq:temperature}
\end{equation}
where $z_i$ is the logit corresponding to the probability at index $i$.

\newcite{sanh2019distilbert} introduced the first distilled Transformer-based language model for English, where the authors created a model that was 1.6x smaller, 2.5x faster, and retained 97\% of the performance of the original BERT \cite{devlin2019bert} on the General Language Understanding Evaluation (GLUE) benchmark \cite{wang2018glue}. Soon after, TinyBERT \cite{jiao2020tinybert} was introduced and it reduced the size of BERT by 7.5x, improved the inference time by 9.4x, while maintained 96.8\% of the performance of the original model on GLUE;  TinyBERT performed distillation at both the pre-training and fine-tuning stages. Also, \newcite{sun2020mobilebert} introduced MobileBERT, a slightly heavier model compared to TinyBERT - 4.3x smaller and 5.5x faster than the original BERT, which managed to retain most of the its knowledge by achieving a GLUE score of 77.7 (0.6 lower than BERT).

One of the first attempts to distill the knowledge of pre-trained language models for languages other than English was BERTino \cite{muffobertino} for Italian. The distillation corpus was composed of 1.8 billion tokens and \textit{bert-base-italian-xxl-uncased}\footnote{\url{https://huggingface.co/dbmdz/bert-base-italian-xxl-uncased}} was used as the teacher model. The resulting model retained most of the knowledge from the original model, as its performance was below with values ranging from 0.29\% to 5.15\% on the evaluated tasks. 

Additionally, a distilled version of Multilingual BERT (mBERT) - LightMBERT - \cite{jiao2021lightmbert} was developed to reduce the discrepancies between languages by transferring the cross-lingual capabilities of the original model into a smaller one. The authors first initialized the student layers with the bottom layers of the teacher - mBERT. Then, they froze the embedding layer, and performed distillation only on the other layers. This simple process allowed them to maintain the cross-lingual generalization capabilities of the original model.

Nevertheless, the knowledge transferred from a single teacher through distillation may be limited and even biased, resulting in a student model of low quality. As such, \newcite{wu2021one} proposed a knowledge distillation approach from several teachers in both the pre-training and fine-tuning stages. Their experimental results showed a significant improvement over single teacher models like DistilBERT or TinyBERT, and even over other models used as teachers.

\section{Distillation Process}

\begin{figure*}
    \centering
    \includegraphics[width=0.95\textwidth]{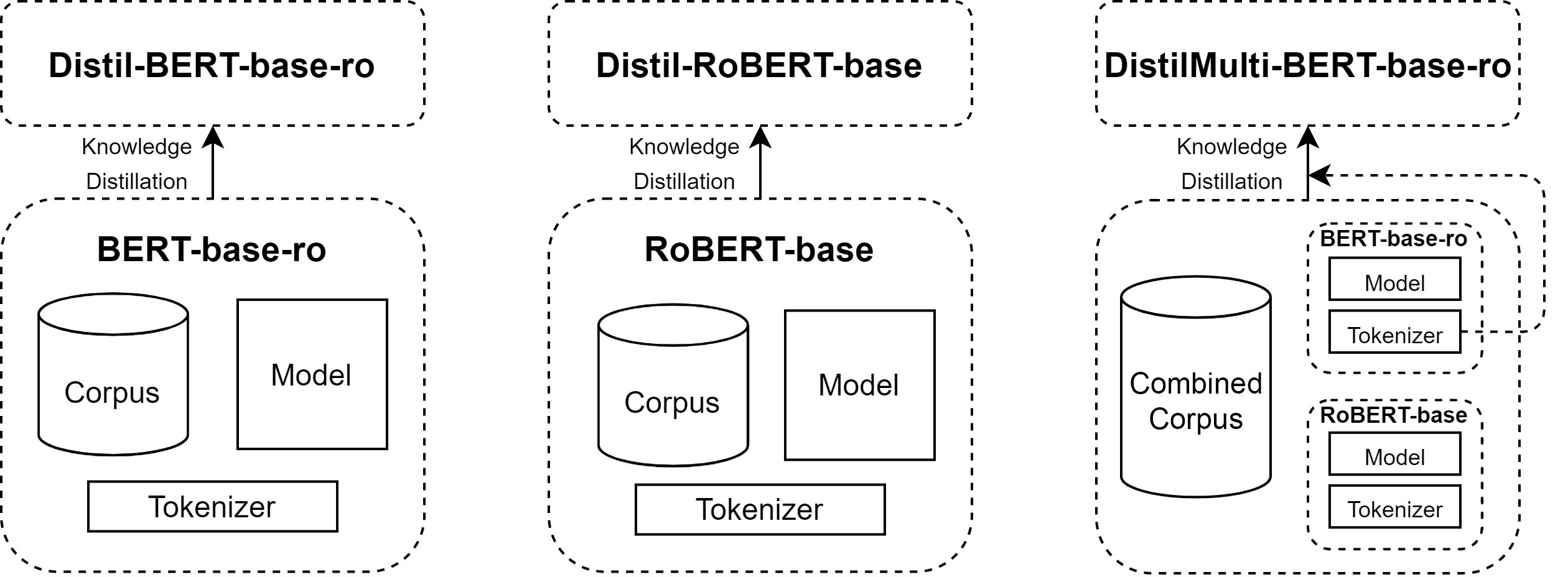}
    \caption{Knowledge distillation process of the three distilled Romanian BERT versions: Distil-BERT-base-ro (left), Distil-RoBERT-base (middle), and DistilMulti-BERT-base-ro (right). DistilMulti-BERT-base-ro uses the BERT-base-ro tokenizer, marked with dotted arrow in the figure.}
    \label{fig:dist}
\end{figure*}

Figure \ref{fig:dist} introduces the overall architecture of the distillation process of all three compressed Romanian BERT variants. The Distil-BERT-base-ro and Distil-RoBERT-base models were obtained by using BERT-base-ro and RoBERT-base as teacher, respectively, together with their teacher's training corpora and tokenizer. DistilMulti-BERT-base-ro used an ensemble of teachers consisting of both Romanian BERTs - BERT-base-ro and RoBERT-base, their combined corpora and the BERT-base-ro tokenizer.

\subsection{Corpora}

\begin{table}[]
    \centering
    \begin{tabular}{lrrr}
         \hline
         \hline
         \textbf{Corpus} & \textbf{Lines} & \textbf{Words} & \textbf{Size} \\
         \hline
         \hline
         RoWiki & 1.5M & 60.5M & 0.4 GB \\
         OPUS & 55.1M & 635M & 3.8 GB \\
         OSCAR & 33.6M & 1725.8M & 11 GB \\
         \hline
         \textbf{Total} & \textbf{90.2M} & \textbf{2421.3M} & \textbf{15.2 GB} \\
         \hline
    \end{tabular}
    \caption{BERT-base-ro training corpus.}
    \label{tab:corpus_bert1}
\end{table}

\begin{table}[]
    \centering
    \begin{tabular}{lrrr}
         \hline
         \hline
         \textbf{Corpus} & \textbf{Lines} & \textbf{Words} & \textbf{Size} \\
         \hline
         \hline
         RoWiki & 2M & 50M & 0.3 GB \\
         RoTex & 14M & 240M & 1.5 GB \\
         OSCAR & 87M & 1780M & 10.8 GB \\
         \hline
         \textbf{Total} & \textbf{103M} & \textbf{2070M} & \textbf{12.6 GB} \\
         \hline
    \end{tabular}
    \caption{RoBERT-base variants training corpus.}
    \label{tab:corpus_bert2}
\end{table}

Tables \ref{tab:corpus_bert1} and \ref{tab:corpus_bert2} summarize the corpora used for training BERT-base-ro and RoBERT-base, respectively. Both training corpora contain the Romanian Wikipedia and the Open Super-large Crawled Aggregated coRpus (OSCAR) \cite{suarez2019asynchronous} datasets, although their final size is not the same due to preprocessing differences. However, the BERT-base-ro corpus uses the OPUS corpus \cite{tiedemann2012parallel} while RoBERT-base uses the RoTex custom collection\footnote{\url{https://github.com/aleris/ReadME-RoTex-Corpus-Builder}}, resulting in a 2.5 GB difference between the final datasets. 

We observed that, in comparison with RoBERT-base training corpus, the BERT-base-ro training corpus was a bit noisier and contained more artifacts; as such, we applied several additional rules to ensure a cleaner version by removing lines in the following conditions: a) lines that contained noise for diacritics (e.g., "\textit{c?nd}" instead of "\textit{când}" - eng., "\textit{when}"); b) lines with uncapitalized versions of the most frequent Romanian named entities (e.g., "\textit{bucurești}" instead of "\textit{București}" - eng.: "\textit{Bucharest}"); and c) lines that were not detected as written in Romanian by \texttt{langdetect}\footnote{\url{https://github.com/Mimino666/langdetect}}. Web page parsing artifacts, such as "\textit{Articolul Anterior}" (eng., "\textit{Previous Article}") at the beginning of the sentence or "Articolul Următor" (eng., "\textit{Next Article}") at the end of the sentence, were also removed. In contrast to the original 15.2 GB of plain text used to train BERT-base-ro, our cleaned version of the corpus contains 14.6 GB of text and approximately 2.3 billions words (i.e., roughly 5.9\% smaller).

The combined corpus used to train DistilMulti-BERT-base-ro was obtained by merging and deduplicating the two corpora, resulting in a corpus that contained 25.3 GB of text and 4.1 billion words words.

\subsection{Teacher Networks}

The teacher models used to perform knowledge distillation during pre-training were the base-cased version of BERT-base-ro \cite{dumitrescu2020birth} for Distil-BERT-base-ro, RoBERT-base \cite{masala2020robert} for Distil-RoBERT-base, and an ensemble of the two for DistilMulti-BERT-base-ro. Although, to the best of our knowledge, two other Romanian BERT models exist, namely RoBERT-small and RoBERT-large that were introduced in the same work as RoBERT-base \cite{masala2020robert}, we only chose these variants because their parameters configuration best fit the compression process used to obtain DistilBERT.

\subsection{Student Networks}

All the distilled versions introduced in this work follow the architecture of DistilBERT, having 6 Transformer encoder layers (i.e., half the layers of BERT-base-ro), 12 attention heads and a hidden dimension of 768. All follow-up analyses consider as reference also mBERT~\cite{devlin2019bert}, the most accessible multi-lingual BERT-based model that supports Romanian; XLM-RoBERTa (XLM-R) \cite{conneau2020unsupervised} was not considered due to its size which is one order of magnitude larger than the other models. Table \ref{tab:model_size} outlines the size, the number of parameters, and several layers statistics (i.e., the number of layers and heads, and the hidden size) for mBERT, RoBERT, BERT-base-ro, and the distilled models. Both Distil-BERT-base-ro and DistilMulti-BERT-base-ro have a size of 312 MB and contain 81 millions parameters, reducing the size of BERT-base-ro by $\sim$35\%. Distil-RoBERT-base is slightly smaller, with 72 millions parameters and a size of 282 MB, compressing RoBERT-base by the same amount\footnote{The difference in the number of parameters between Distil-BERT-base-ro and RoBERT-base is generated by the difference in vocabulary size.}.

\begin{table*}
    \centering
    \begin{tabular}{lrrrrrr}
         \hline
         \hline
         \textbf{Model} & \textbf{Layers} & \textbf{Hidden} & \textbf{Heads} & \textbf{Vocab} & \textbf{Size} & \textbf{Params} \\
         \hline
         \hline
         mBERT & 12 & 768 & 12 & 120K & 681 MB & 177M \\
         BERT-base-ro & 12 & 768 & 12 & 50K & 477 MB & 124M \\
         RoBERT-small & 12 & 256 & 8 & 38K & 74 MB & 19M \\
         RoBERT-base & 12 & 768 & 12 & 38K & 441 MB & 114M \\
         RoBERT-large & 24 & 1024 & 16 & 38K & 1.3 GB & 341M \\
         \hline
         Distil-BERT-base-ro & 6 & 768 & 12 & 50K & 312 MB & 81M \\
         Distil-RoBERT-base & 6 & 768 & 12 & 38K & 282 MB & 72M \\
         DistilMulti-BERT-base-ro & 6 & 768 & 12 & 50K & 312 MB & 81M \\
         \hline
    \end{tabular}
    \caption{Model size comparison of mBERT, Romanian BERTs, and our distilled versions.}
    \label{tab:model_size}
\end{table*}

Following \newcite{muffobertino}, the total loss function $\mathcal{L}$ used to pre-train Distil-BERT-base-ro and Distil-RoBERT-base was composed of three parts $\mathcal{L}_{KD}$, $\mathcal{L}_{MLM}$ and $\mathcal{L}_{COS}$, weighted by different coefficients that sum to one: \begin{equation}
    \mathcal{L} = \lambda_{KD}\mathcal{L}_{KD} + \lambda_{MLM}\mathcal{L}_{MLM} + \lambda_{COS}\mathcal{L}_{COS}
\end{equation}
where $\mathcal{L}_{KD}$ is the knowledge distillation loss \cite{44873}, $\mathcal{L}_{MLM}$ is the masked language modeling (MLM) loss \cite{devlin2019bert}, $\mathcal{L}_{COS}$ is the cosine similarity embedding loss used to align the hidden states of the teacher and student models \cite{sanh2019distilbert}, and $\lambda_{KD}$, $\lambda_{MLM}$ and $\lambda_{COS}$ are the weights of each loss. A higher weight for the knowledge distillation loss $\lambda_{KD}=0.625$ is considered because the training corpora for the student network represents a large portion ($\sim$94\%) of the pre-training corpora for the teacher; as such, more of the internal knowledge acquired by the teachers would be transferred to the students in the process, at the detriment of not learning much by itself based on the MLM loss. $\lambda_{MLM}$ and $\lambda_{COS}$ are set to 0.25 and 0.125, respectively.

The knowledge distillation loss $\mathcal{L}_{KD}$ and the cosine similarity loss $\mathcal{L}_{COS}$ were split for training DistilMulti-BERT-base-ro into two equally weighted parts corresponding to the losses of each model in the ensemble :

\begin{equation}
    \mathcal{L}_{KD} = \frac{\mathcal{L}^1_{KD} + \mathcal{L}^2_{KD}}{2}
\end{equation}

\begin{equation}
    \mathcal{L}_{COS} = \frac{\mathcal{L}^1_{COS} + \mathcal{L}^2_{COS}}{2}
\end{equation}

where $\mathcal{L}^1_{KD}$, $\mathcal{L}^1_{COS}$ are the distillation and cosine similarity losses corresponding to BERT-base-ro, and $\mathcal{L}^2_{KD}$, $\mathcal{L}^2_{COS}$ are the distillation and cosine similarity losses corresponding to RoBERT-base.

\subsection{Training Settings} An important practical aspect of knowledge distillation is the initialization of the student parameters. Following the setup of DistilBERT, we initialize the students layers with the teacher layers by taking out the first layer of two\footnote{We initialized DistilMulti-BERT-base-ro with the parameters of BERT-base-ro.} (i.e. first layer, third layer, fifth layer etc.). The distilled models were trained for 3 epochs with a batch size of 256 and a learning rate of 5e-4. We also applied a weight decay of 1e-4 and a warm-up for the first 5\% of the entire training process, together with gradient clipping \cite{pascanu2013difficulty} for gradient norms surpassing the value of 5. The temperature $T$ from Equation \ref{eq:temperature} was set to 2. The training process took approximately 30 days on two GeForce GTX 1080 Ti for each model variant.

\begin{table}
    \centering
    \begin{tabular}{lcccc}
         \hline
         \hline
         \textbf{Task} & \textbf{Epochs} & \textbf{Batch} & \textbf{Warm} & \textbf{L-Rate} \\
         \hline
         \hline
         UPOS & 10 & 16 & 1000 & 1e-4 \\
         XPOS & 10 & 16 & 1000 & 4e-5 \\
         NER & 15 & 16 & 500 & 5e-5 \\
         SAPN & 10 & 16 & 1000 & 3e-5 \\
         SAR & 10 & 16 & 1000 & 5e-5 \\
         DI & 5 & 8 & 1500 & 5e-5 \\
         STS & - & 256 & 0 & 2e-5 \\
         \hline
    \end{tabular}
    \caption{Hyperparamters used to fine-tune the models on each evaluation task.}
    \label{tab:hyperparam}
\end{table}

\section{Task Evaluation}

Five Romanian tasks were considered to create a strong evaluation setup for our distilled models:

\begin{itemize}
    \item \textbf{Part-of-Speech (POS) Tagging}: Label a sequence of tokens with Universal Part-of-Speech (UPOS) and eXtended Part-of-Speech (XPOS).
    \item \textbf{Named Entity Recognition (NER)}: Tag a sequence of tokens with Inside–Outside–Beginning (IOB) labels \cite{ramshaw1999text}.
    \item \textbf{Sentiment Analysis (SA)}: Predict whether a review expresses a positive or negative sentiment (SAPN), together with its rating (SAR).
    \item \textbf{Dialect Identification (DI)}: Identify the Romanian/Moldavian dialects in news articles.
    \item \textbf{Semantic Textual Similarity (STS)}: Given a pair of sentences, measure how semantically similar they are.
\end{itemize}

The fine-tuning of the models for each task was performed by using the AdamW optimizer \cite{loshchilov2018decoupled} and a scheduler that linearly decreased the learning rate to 0 at the end of the training. The number of epochs, the batch size, the learning rate, and the warm-up steps used for each individual task are depicted in Table \ref{tab:hyperparam}. In line with the recommendations of the authors of the STS evaluation scripts\footnote{\url{https://github.com/dumitrescustefan/RO-STS/tree/master/baseline-models}}, we did not employ any warm-up steps, a higher batch size of 256 was considered, and early stopping was used for training instead of a maximum number of epochs. Each experiment was run 5 times and we report the average scores in order to mitigate the variation in performance due to the random initialization of the weights.

\subsection{Part-of-Speech Tagging}

\paragraph{Task Description.} The original splits of the Romanian Reference Trees (RRT) corpus \cite{barbu2016romanian} from Universal Dependencies (UD) v2.7 were used to train and evaluate the distilled models on UPOS and XPOS tagging. The evaluation metric employed to measure the performance on both subtasks was the macro-averaged F1-score.

\paragraph{Methodology.} The output embeddings of the model $E_i$ were projected into a tensor of dimension equal with the number of classes of UPOSes or XPOSes. The transformation function was a feed-forward layer with weights $W$, bias $b$, and the $LeakyReLU$ activation function that produced the logits $y_i$ corresponding to each embedding (i.e., $y_i = LeakyReLU(W^TE_i + b)$); a dropout of 0.1 was also applied for regularization. The logits were further transformed using the $softmax$ function to obtain the output distribution. The cross-entropy loss between the target labels and the predicted probabilities was employed as the value to be minimized.

\paragraph{Results} Model performance on UPOS and XPOS prediction are outlined in Table \ref{tab:upos_xpos_eval}. The highest F1-score of 98.07\% on UPOS evaluation was achieved by DistilMulti-BERT-base-ro, outperforming both its teachers and being the second model in the overall leaderboard behind RoBERT-large. The best XPOS performance was obtained by Distil-BERT-base-ro with 97.08\%, surpassing BERT-base-ro with 96.46\%, but falling behind its teacher and the large BERT variant. 

An in-depth analysis of Romanian UPOS and XPOS evaluation was performed in \cite{Pais2021romjist}, where several basic language processing kits (BLARK) were tested on RRT. Our distilled models obtained superior results on UPOS and comparable results with the best models on XPOS.

\begin{table}
    \centering
    \begin{tabular}{lcc}
        \hline
        \hline
        \textbf{Model} & \textbf{UPOS}  & \textbf{XPOS} \\
        \hline
        \hline
        mBERT & 97.87 & 96.16 \\
        RoBERT-small & 97.43 & 96.05 \\
        RoBERT-base & 98.02 & 97.15 \\
        RoBERT-large & \textbf{98.12} & \textbf{97.81} \\
        BERT-base-ro & 98.00 & 96.46 \\
        \hline
        Distil-BERT-base-ro & 97.97 & 97.08 \\
        Distil-RoBERT-base & 97.12 & 95.79 \\
        DistilMulti-BERT-base-ro & 98.07 & 96.83 \\
        \hline
    \end{tabular}
    \caption{UPOS and XPOS evaluation results on RRT.}
    \label{tab:upos_xpos_eval}
\end{table}

\subsection{Named Entity Recognition}

\paragraph{Task Description.} NER evaluation was performed on Romanian Named Entity Corpus (RONEC) \cite{dumitrescu2020introducing}. Models were evaluated according to \newcite{segura2013semeval} and the macro-averaged F1-scores are reported for the exact matches of the IOB labels.

\paragraph{Methodology.} The approach used to fine-tune the models is the same as the one used in POS tagging, the only difference being the dimension of the output tensor that was adjusted to match the number of classes found in RONEC.

\paragraph{Results} The results on this task are presented in Table \ref{tab:ronec_metrics}. The Distil-BERT-base-ro obtained a strict F1-score of 79.42\% which is almost identical to the strict F1-score obtained by the distilled ensemble - 79.43\%, both models outperforming Distil-RoBERT-base by approximately 0.3\% on this metric. Our compressed models lagged behind the base models by more than 3\% on the strict F1 metric (i.e., RoBERT-large and BERT-base-ro) and by more than 2.5\% on the exact F1-score (i.e., BERT-base-ro), but they managed to achieve a performance close to the teachers on the other two metrics.

\setlength{\tabcolsep}{4pt}
\begin{table*}
    \centering
        \begin{tabular}{lcccc}
              \hline
              \hline
              \textbf{Model} & \textbf{Type} & \textbf{Partial} & \textbf{Strict} & \textbf{Exact} \\
              \hline
              \hline
                mBERT & 84.52 & 86.27 & 80.60 & 84.13 \\
                RoBERT-small & 83.11 & 84.59 & 78.61 & 82.06\\
                RoBERT-base & 85.92 & 87.21 & 82.05 & 85.14 \\
                RoBERT-large & \textbf{86.45} & 87.19 & \textbf{82.61} & 85.09 \\
                BERT-base-ro & 86.21 & \textbf{87.84} & 82.54 & \textbf{85.88} \\ 
                \hline
                Distil-BERT-base-ro & 83.83 & 85.73 & 79.42 & 83.35 \\
                Distil-RoBERT-base & 85.80 & 87.39 & 79.15 & 83.11 \\
                DistilMulti-BERT-base-ro & 85.48 & 87.66 & 79.43 & 83.22 \\
              \hline
        \end{tabular}
    \caption{NER evaluation results on RONEC.}
    \label{tab:ronec_metrics}
\end{table*}

\subsection{Sentiment Analysis}

\paragraph{Task Description}
SA was performed on the Large Romanian Sentiment Data Set (LaRoSeDa) \cite{tache2021clustering}, a dataset that contains 15,000 reviews written in Romanian, of which 7,500 are positive and 7,500 negative. We fine-tuned the models to predict both positive and negative sentiments, as well as the rating of each review. The models were evaluated by measuring the accuracy and the F1-macro score for the two prediction subtasks.

\paragraph{Methodology}
The model was given as input two sentences - the title of the review and the content of the review - separated by the token \texttt{[SEP]}. We projected the embedding  $C$ of the token \texttt{[CLS]} into a scalar $y$, representing the sentiment of the review, by using a linear layer with weights $W$ and bias $b$, to which the $LeakyReLU$ activation function is applied (i.e., $y =\sigma(W^TC + b)$). Then, the $sigmoid$ function is applied to the scalar $y$ to obtain the output probability. 

For the rating prediction, the input was also composed of the title and the content of the review separated by the \texttt{[SEP]} token. However, the embedding corresponding to the \texttt{[CLS]} token was projected using a feed-forward neural network into four dimensions representing the logits of possible ratings. Also, the $softmax$ function is employed instead of the $sigmoid$ function to transform the output of the neural network. The models were trained to reduce the binary cross-entropy for SAPN and the cross-entropy losses for SAR, respectively.

\paragraph{Results}

Table \ref{tab:sa_results} depicts the results of the SA task. DistilRo-BERT achieved the best performance on the positive versus negative sentiment analysis out of all other distilled models, with 98.20\% accuracy and a 98.12\% F1-score. For rating prediction, DistilRo-BERT obtained an accuracy score of 90.14\% and a F1-score of 80.51\%\, surpassing all the other evaluated models. The difference in scores between the sentiment and the rating prediction for each model might be due to the noisy labeling of the ratings, as stated by the authors of the dataset \cite{tache2021clustering}.

\setlength{\tabcolsep}{5.5pt}
\begin{table*}
    \centering
        \begin{tabular}{lcccc}
            \hline
            \hline
             & \multicolumn{2}{c}{\textbf{\underline{SAPN}}} & \multicolumn{2}{c}{\textbf{\underline{SAR}}} \\
             \textbf{Model} & \textbf{Acc} & \textbf{F1} & \textbf{Acc} & \textbf{F1} \\ 
             \hline
             \hline
             mBERT & 97.43 & 97.28 & 88.12 & 78.98 \\
             RoBERT-small & 97.37 & 97.22 & 87.55 & 77.81\\
             RoBERT-base & \textbf{98.30} & \textbf{98.20} & 89.69 & 79.40 \\
             RoBERT-large & 98.25 & 98.16 & 89.91 & 79.82 \\
             BERT-base-ro & 98.07 & 97.94 & 88.45 & 79.61 \\ 
             \hline
             Distil-BERT-base-ro & 98.20 & 98.12 & \textbf{90.14} & \textbf{80.51} \\
             Distil-RoBERT-base & 98.01 & 97.61 & 88.33 & 79.58 \\
             DistilMulti-BERT-base-ro & 98.11 & 97.74 & 89.43 & 79.77 \\
             \hline
        \end{tabular}
    \caption{SA evaluation results for positive and negative and rating classification on LaRoSeDa.}
    \label{tab:sa_results}
\end{table*}

\subsection{Dialect Identification}

\paragraph{Task Description} 
The MOldavian and ROmanian Dialectal COrpus (MOROCO) \cite{butnaru2019moroco} is a dataset that contains 33,564 samples of text that were collected from news, annotated with their dialect and with one of six topics that each sample belongs to. We evaluated the proposed models in terms of accuracy and F1-score (macro averaged) on the identification of Romanian versus Moldavian dialects on MOROCO samples.

\paragraph{Methodology} The same approach as in the previously described positive versus negative sentiment analysis being a binary classification task was considered for this task. The only slight difference is that the MOROCO samples do not have a title; as such, no separation of the title and the content was necessary.

\paragraph{Results} DI results are outlined in Table \ref{tab:di_results}. As it can be observed, Distil-BERT-base-ro and DistilMulti-BERT-base-ro achieve better performance than their teachers on this task (i.e., accuracy/F1-scores of 96.36\%/96.31\% and 96.26\%/96.16\%), outperforming even the large variant of RoBERT.

\begin{table}
    \centering
    \begin{tabular}{lcc}
        \hline
        \hline
        \textbf{Model} & \textbf{Acc} & \textbf{F1} \\
        \hline
        \hline
        mBERT & 95.12 & 95.06 \\
        RoBERT-small & 95.48 & 95.41 \\
        RoBERT-base & 96.10 & 96.07 \\
        RoBERT-large & 96.20 & 96.17 \\
        BERT-base-ro & 95.64 & 95.58 \\
        \hline
        Distil-BERT-base-ro & \textbf{96.36} & \textbf{96.31} \\
        Distil-RoBERT-base & 96.13 & 96.11 \\
        DistilMulti-BERT-base-ro & 96.26 & 96.18 \\
        \hline
    \end{tabular}
    \caption{DI evaluation results on MOROCO.}
    \label{tab:di_results}
\end{table}

\subsection{Semantic Textual Similarity}

\paragraph{Task Description} Romanian Semantic Textual Similarity (RoSTS)\footnote{\url{https://github.com/dumitrescustefan/RO-STS}} is a dataset that was obtained by translating the Semantic Textual Similarity (STS) dataset\footnote{\url{https://ixa2.si.ehu.eus/stswiki/index.php/STSbenchmark}} and used to compare the capacity of the employed BERT models in capturing the semantic similarity between Romanian sentences. The evaluation metrics are the Pearson and the Spearman coefficients.

\paragraph{Methodology} The fine-tuning of the language models was performed by giving as input the two sentences separated by the \texttt{[SEP]} token, followed by the projection of the resulted \texttt{[CLS]} token embedding into a scalar $y$ using a linear layer with weights $W$ and bias $b$, to which the $LeakyRELU$ function is applied (i.e., $y = \sigma(W^TC + b)$). Then, the $sigmoid$ function is applied to the output logit. To match with the $y$ interval, the original similarities were normalized from the $[0,5]$ to $[0,1]$. The models were trained using the mean squared error loss.

\paragraph{Results} Task results are outlined in Table \ref{tab:sts_results}. DistilMulti-BERT-base-ro achieved the highest average Pearson/Spearman scores (80.66\%/80.27\%) out of all distilled models, slightly outperforming Distil-BERT-base-ro by approximately 0.1\%/0.25\% and Distil-RoBERT-base by approximately 0.85\%/0.25\%. In comparison with the teacher models, both Distil-BERT-base-ro and DistilMulti-BERT-base-ro achieved better results than BERT-base-ro, but lagged behind RoBERT-base by approximately 0.5\% on both metrics and RoBERT-large by approximately 2\%/1.5\%.

\begin{table}
    \centering
    \begin{tabular}{lcc}
        \hline
        \hline
        \textbf{Model} & \textbf{Pears} & \textbf{Spear} \\
        \hline
        \hline
        mBERT & 76.64 & 76.41 \\
        RoBERT-small & 78.24 & 77.84 \\
        RoBERT-base & 81.18 & 80.69 \\
        RoBERT-large & \textbf{82.51} & \textbf{81.83} \\
        BERT-base-ro & 80.30 & 79.94 \\
        \hline
        Distil-BERT-base-ro & 80.57 & 80.02 \\
        Distil-RoBERT-base & 79.80 & 78.82 \\
        DistilMulti-BERT-base-ro & 80.66 & 80.27 \\
        \hline
    \end{tabular}
    \caption{STS evaluation results on RoSTS.}
    \label{tab:sts_results}
\end{table}

\section{Loyalty}

\newcite{xu2021beyond} introduced two new metrics - loyalty and robustness - for measuring the similarities between a student and a teacher model because just comparing the evaluation metric for a specific task does not reflect how alike the student and the teacher model behave. For loyalty, the authors measure similarities between the labels (L-L) and the probabilities (P-L) predicted by the models fine-tuned on Multi-Genre Natural Language Inference dataset \cite{williams2018broad}. However, we choose to evaluate the loyalties of our distilled models on the test set of LaRoSeDa as there is no natural language inference dataset for Romanian. In addition, we introduce a new loyalty evaluation metric for measuring the similarity for regression - regression loyalty (R-L) -, using the test set of RoSTS as target. Robustness was measured by \newcite{jin2020bert} using the after-attack accuracy (AA); however, it is not used in this work due to the lack of high-quality word embeddings that are retrofitted for synonymy\footnote{It must be noted that a version of synonymy aware word embeddings was introduced in \cite{dumitrescu2018rowordnet}, but they were not open-sourced for public usage.}. The loyalty metrics for MultiDistil-BERT-base-ro were computed using the average of the metrics computed with each individual teacher.

\begin{table}
    \centering
        \begin{tabular}{lccc}
             \hline
             \hline
             \textbf{Model} & \textbf{L-L} & \textbf{P-L} & \textbf{R-L} \\
             \hline
             \hline
             Distil-BERT-base-ro & 87.80 & \textbf{74.76} & 94.05 \\
             Distil-RoBERT-base & 84.63 & 71.95 & 92.24 \\
             DistilMulti-BERT-base-ro & \textbf{89.75} & 73.23 & \textbf{94.64} \\
             \hline
        \end{tabular}
    \caption{Label, probability and regression loyalty results between our distilled models and their teachers.}
    \label{tab:loyal_results}
\end{table}

\subsection{Label Loyalty}

Label Loyalty (L-L) directly measures the similarity between the labels predicted by two models as the accuracy between the teacher labels acting as the ground truth versus the student labels:

\begin{equation}
    LL = Accuracy(label_t, label_s)
\end{equation}
where $label_t$ and $label_s$ are the labels predicted by the teacher and the student, respectively.

The highest L-L score was obtained by DistilMulti-BERT-base-ro with 89.75\%, as outlined in Table \ref{tab:loyal_results}, showing that the distilled ensemble has superior label alignment with its teachers than each individual distilled model.

\subsection{Probability Loyalty}

Computing the similarities between the output probabilities of a teacher and student after compression is also important in industrial applications that focus on confidence and meaningfulness \cite{guo2017calibration}. P-L evaluates the distance between the teacher output probabilities $P_t$ and the student output probabilities $P_s$ as:

\begin{equation}
    PL = 1 - \sqrt{D_{JS}(P_t || P_s)}
\end{equation}

where $D_{JS}$ is the Jensen–Shannon divergence, defined using the Kullback–Leibler divergence over probabilities $X$ $D_{KL}(P||Q) = \sum_{x \in X}P(x)log(\frac{P(x)}{Q(x)})$ as:

\begin{equation}
    D_{JS}(P_t || P_s) = \frac{D_{KL}(P_t || P_s) + D_{KL}(P_s || P_t)}{2}
\end{equation}

The results of P-L evaluation are depicted in the second column of Table \ref{tab:loyal_results}. The highest score was obtained by Distil-BERT-base-ro with 74.76\%, outperforming the distilled ensemble model with approximately 1.5\%. 

\subsection{Regression Loyalty}

The previous two metrics do not assess performance in the case of regression problems. Thus, this work introduces regression loyalty as the Pearson Correlation Coefficient \cite{breese1998empirical} between the output of the teacher $pred_t$ and the output of the student $pred_s$:

\begin{equation}
    RL = Pearson(pred_t, pred_s)
\end{equation}

The results for R-L are outlined in the last column of Table \ref{tab:loyal_results}. DistilMulti-BERT-base-ro achieved the highest regression loyalty with its two teachers (a 94.64\% R-L score), followed by Distil-BERT-base-ro (94.05\%) and Distil-RoBERT-base (92.24\%).


\section{Inference Speed}

\begin{figure}
    \centering
    \includegraphics[width=0.52\textwidth]{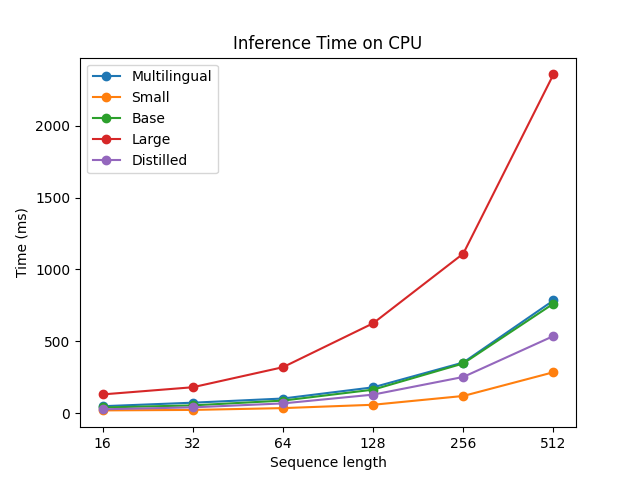}
    \includegraphics[width=0.52\textwidth]{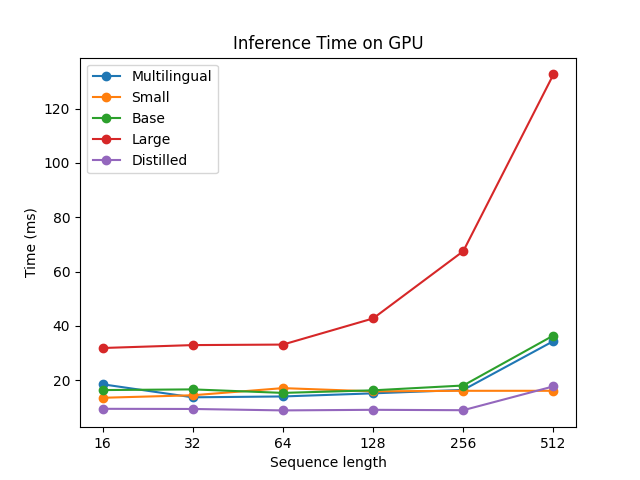}
    \caption{Inference time on CPU (up) and on GPU (down) of the evaluated models, grouped by their dimensions: Multilingual (mBERT), Small (RoBERT-small), Base (BERT-base-ro, RoBERT-base), Large (RoBERT-large) and Distilled (Distill-BERT-base-ro, Distill-RoBERT-base, DistillMulti-BERT-base-ro).}
    \label{fig:speed}
\end{figure}

The inference time of the three distilled versions in comparison with the other Romanian or multilingual models was evaluated on random sequences with lengths ranging from 16 tokens to 512 tokens, using both a CPU - Intel i7-7700K - and a GPU - GeForce GTX 1080 Ti. The models were grouped by their size into four categories to make the visualization more appealing, namely: Multilingual (i.e., mBERT), Small (i.e., RoBERT-small), Base (i.e., BERT-base-ro and RoBERT-base), Large (i.e., RoBERT-large), and Distilled (i.e., Distill-BERT-base-ro, Distill-RoBERT-base, and DistillMulti-BERT-base-ro). The inference times are depicted in Figure \ref{fig:speed}. 

The distilled models obtained a significant improvement on the GPU, being almost twice as fast as the base and small models, and at least three times faster than the RoBERT-large. It should be noted that the small variant obtains a similar speed on the GPU for larger sequence lengths due to more parallelization happening at the level of the input tokens; RoBERT-small benefits from its reduced size and overcomes its disadvantage in parallelization by having 12 layers instead of 6. The inference times on the CPU of the distilled models is also better than those obtained by the base and large models, but worse than those of RoBERT-small because it has less parameters. 

\section{Conclusions}

Creating models for low resource languages is an important area of research, aiming to empower the usage of NLP in various contexts. This work introduces three distilled models from Romanian BERT models: Distil-BERT-base-ro, Distil-RoBERT-base, and DistilMulti-BERT-base-ro, that are ~35\% smaller than their teacher models (BERT-base-ro and RoBERT-base). Each model was evaluated on five Romanian NLP tasks: part-of-speech tagging, named entity recognition, sentiment analysis, dialect identification, and semantic textual similarity. Our experimental results showed that the distilled models maintain most of the prediction performance of their original models, while containing only half of their layers. Moreover, the distilled models even outperformed their teachers and other evaluated models on several tasks. We further tested the inference speed of the three distilled models on various sequence lengths, and our results outlined a reduction of almost 50\% on a single GPU in comparison to their teacher models.

Future work considers the creation of a distilled version using RoBERT-large as teacher, as well as further reducing the size of the current models by pruning and/or quantizing their weights. In addition, we intend to fine-tune the distilled models and annotate the  Romanian sub-corpus from the CURLICAT multilingual corpus covering domains relevant for CEF Digital Service Infra-structures (DSIs). Besides standard mark-up (lemmatization, POS tagging, chunking, dependency parsing)  annotation will include recognition and labeling of the occurrences of terms recorded in the Interactive Terminology for Europe (IATE) \footnote{\url{https://iate.europa.eu/home}}, s described in the Curated Multilingual Language Resources for CEF.AT (CURLICAT) \footnote{\url{https://curlicat.eu/}} project. Finally, the results presented in this paper are going to be included on the five evaluation tasks in LiRo \cite{dumitrescu2021liro}, a benchmark for Romanian language NLP tasks.

\section*{Acknowledgements}

This research was supported by the EC grant INEA/CEF/ICT/A2018/28592472 for the Action No: 2019-EU-IA-0034 entitled “Curated Multilingual Language Resources for CEF.AT” (CURLICAT) and by the Romanian Ministry of European Investments and Projects through the Competitiveness Operational Program (POC) project "HOLOTRAIN" (grant no. 29/221, SMIS code: 129077).

\section{Bibliographical References}\label{reference}

\bibliographystyle{lrec2022-bib}
\bibliography{lrec2022-example}

\end{document}